\documentclass[sn-nature]{sn-jnl}

\usepackage{graphicx}
\usepackage{multirow}
\usepackage{amsmath,amssymb,amsfonts}
\usepackage{amsthm}
\usepackage{mathrsfs}
\usepackage[title]{appendix}
\usepackage{xcolor}
\usepackage{textcomp}
\usepackage{manyfoot}
\usepackage{booktabs}
\usepackage{longtable}
\usepackage{algorithm}
\usepackage{algorithmicx}
\usepackage{algpseudocode}
\usepackage{listings}
\usepackage{xspace}
\usepackage{lmodern}
\usepackage{soul}
\usepackage{makecell}
\usepackage{tcolorbox}
\usepackage{csquotes}

\usepackage{ifthen}

\newcommand{\ourmethod}{Survprompt\xspace}

\newif\ifdraft
\draftfalse

\ifdraft
    \SetWatermarkText{Confidential working draft, please do not share contents or document without permission}
    \SetWatermarkScale{1}
    \SetWatermarkColor[HTML]{B22222}
    \SetWatermarkAngle{0}
    \SetWatermarkFontSize{12pt}
    \SetWatermarkHorCenter{0.5\paperwidth}
    \SetWatermarkVerCenter{0.9\paperheight}
\fi

\begin{document}

\title[Article Title]{Large Language Models are Approximate Survival Estimators}

\author*[1]{\fnm{Juan M} \sur{Zambrano Chaves}}\email{juanza@microsoft.com}
\equalcont{Equal contribution}
\author*[1]{\fnm{Peniel} \sur{Argaw}}\email{penielargaw@microsoft.com}
\equalcont{Equal contribution}
\author[1]{\fnm{Risa} \sur{Ueno}}
\equalcont{Equal contribution}
\author[2,3]{\fnm{Carlo} \sur{Bifulco}}
\author[3,4]{\fnm{Kristina} \sur{Young}}
\author[3]{\fnm{Rom} \sur{Leidner}}
\author[1]{\fnm{Tristan} \sur{Naumann}}
\author[1]{\fnm{Hoifung} \sur{Poon}}

\affil[1]{Microsoft Research}
\affil[2]{Providence Genomics, Portland, OR, USA}
\affil[3]{Earle A. Chiles Research Institute, Providence Cancer Institute, Portland, OR, USA}
\affil[4]{The Oregon Clinic, Radiation Oncology Division, Portland, OR, USA}

\abstract{
Survival analysis estimates time-to-event outcomes from patient covariates and is widely used for risk assessment in medical applications.
Patients seeking prognostic information following a diagnosis may turn to large language models (LLMs), which are now readily accessible through consumer applications.
However, despite LLMs' success in various biomedical tasks, whether they can provide accurate survival predictions has not been rigorously evaluated.
Here, we introduce \ourmethod, a framework that converts structured patient covariates into free-text clinical vignettes and prompts pre-trained LLMs to predict survival in a zero-shot fashion.
We benchmark \ourmethod against conventional survival models, such as random survival forests (RSF), across two multi-institutional pan-cancer cohorts: the publicly available MSK-CHORD cohort and a newly curated cohort constructed from the Providence St. Joseph Health Network via an LLM-based medical abstraction framework.
We report censored mean absolute error (cMAE) and concordance index (c-index), and conduct feature ablation analysis to identify clinical variables influencing LLM predictions.
Frontier LLMs achieved surprisingly competitive cMAE for individual survival times in zero-shot fashion.
For example, GPT-5.6-Sol achieved a cMAE within 10\% of state-of-the-art RSF models specifically trained for survival prediction for several cancer types, and a lower cMAE than RSF for patients with prostate cancer in MSK-CHORD. Feature ablation revealed that LLMs prioritized clinical variables in a manner analogous to specialized survival models. However, this aggregate performance revealed critical limitations: LLMs showed inconsistent accuracy across cancer types and institutions and demonstrated poor ability to discriminate between high- and low-risk patients (lower c-index).
Zero-shot LLMs can generate surprisingly accurate prognostic estimates without specialized training, but their variable performance across cancer types and institutions presents an important limitation for clinical use.
Clinicians and patients who consult LLMs for prognosis should be aware of these limitations before acting on their outputs.
}

\maketitle

Cancer remains a leading cause of mortality worldwide, making accurate survival prediction critical for guiding treatment decisions, determining clinical trial eligibility, and improving patient outcomes.
Traditional survival analysis methods, such as Cox proportional hazards models, random survival forests (RSFs), and shallow neural networks, are effective and widely used, providing valuable insights into risk factors and survival probabilities \cite{coxph, rsf, deeplearnsurvrev}.
However, these approaches have inherent limitations, particularly in their reliance on manually engineered features, rigid assumptions about covariate relationships, and the need for dataset-specific training.

A key challenge with traditional survival analysis methods is their dependence on structured tabular data, which requires expert-defined features and often requires either imputation of missing values or exclusion of individuals with missing data from analyses \cite{carroll2020missing}.
Additionally, such methods often assume predefined functional relationships, such as proportional hazards or additive effects, between covariates and survival time, which can restrict their ability to model complex, nonlinear interactions \cite{wongvibulsin2019rfslam}.
Furthermore, these models typically need to be trained from scratch for each new dataset, making them data-intensive and difficult to adapt across clinical settings \cite{gupta2020transfer}.
These challenges highlight the need for survival analysis methods that enable individualized survival estimates, generalize across datasets with minimal retraining, and offer greater flexibility in handling patient data and modeling of input features.

Large language models (LLMs) offer a potential solution to these challenges.
Unlike traditional survival models, LLMs can flexibly process a wide range of inputs, from structured tabular data to free-text vignettes, eliminating the need for rigid feature engineering \cite{tabllm}.
They can capture complex, nonlinear relationships between variables, leveraging vast pre-trained knowledge to make predictions with minimal labeled data \cite{engels2025not, incontextlearning}.
Additionally, LLMs are highly adaptable, capable of zero- and few-shot learning, allowing them to generalize across different datasets without the need for retraining \cite{llmfewshot, llms0shot}.

However, LLMs have historically struggled with mathematical reasoning \cite{llmsmath, llmsmathreason}, and remain largely underexplored as outcome estimators in clinical scenarios despite recent work showing they can be surprisingly effective at regression tasks \cite{txllm, tang2025understanding, song2025decodingbased}.
Importantly, patients frequently seek prognostic information following a cancer diagnosis, and with LLMs now readily accessible through consumer applications, they may turn to these tools for survival estimates.
Whether LLMs can provide reliable answers to such queries remains unknown and warrants systematic evaluation.

In this work, we introduce \ourmethod, a framework that systematically evaluates LLMs for zero-shot direct cancer survival prediction.
Our contributions are as follows: (i) we curate and harmonize two large-scale multi-institutional pan-cancer datasets spanning five cancer types each, namely the publicly available MSK-CHORD cohort \cite{mskchord} and a newly constructed cohort built via a LLM-based medical abstraction framework \cite{UMA}; (ii) we develop a method to convert structured tabular covariates into free-text clinical vignettes, and propose \ourmethod, a zero-shot inference framework that prompts pre-trained LLMs to predict survival from these vignettes; (iii) we benchmark \ourmethod against conventional survival models, demonstrating competitive censored mean absolute error alongside limitations in individual risk discrimination; and (iv) we use feature ablation analysis and time-dependent error visualizations to characterize which clinical variables influence LLM predictions and where LLM-based prediction succeeds or fails.

We evaluate \ourmethod against conventional survival models across multiple cancer types and institutions.
We find that LLMs like GPT-5.6-Sol can achieve surprisingly competitive performance in estimating individual survival times (censored mean absolute error), addressing the specific question patients ask: "how long do I have?"
However, our comprehensive evaluation also reveals critical limitations: LLMs demonstrate inconsistent performance across cancer types and institutions and show poor ability to discriminate between high- and low-risk patients.
To our knowledge, these findings provide the first rigorous zero-shot assessment of frontier LLMs for cancer survival prediction spanning multiple cancer types and institutions. Furthermore, we address the rapidly emerging reality that patients are increasingly relying on AI systems as a source of medical guidance.

\section*{Results}\label{results}
\subsection*{Overview of \ourmethod}
\begin{figure}[h]
\centering
\includegraphics[width=0.9\textwidth]{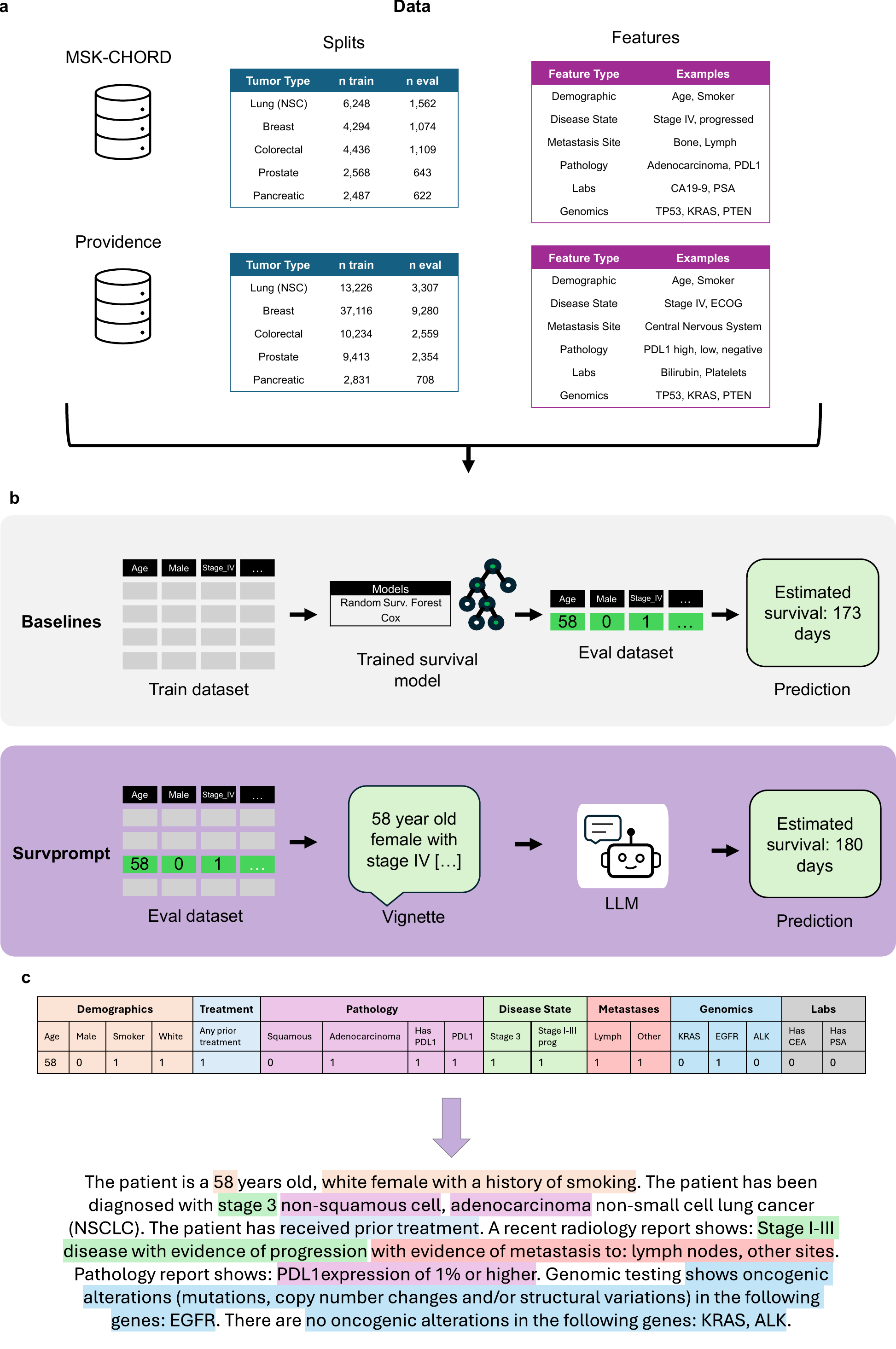}
\caption{\textbf{Overview of \ourmethod}. \textbf{a,} Two multi-cancer datasets (MSK-CHORD and Providence) are used for model development and validation. \textbf{b,} Baseline survival models representing conventional approaches require cancer-type and dataset-specific training on tabular clinical features, whereas \ourmethod converts structured data into free-text clinical vignettes for direct input into a large language model (LLM) to estimate patient survival. \textbf{c,} Construction of free-text clinical vignettes from structured tabular features. NSCLC, non-small cell lung cancer; LLM, large language model}\label{overview}
\end{figure}

We evaluated \ourmethod against cancer- and dataset-specific RSF baselines trained on the same structured features, across the MSK-CHORD and Providence cohorts of patients with five cancer types: non-small cell lung cancer (NSCLC), breast (BRCA), colorectal (CRC), pancreatic (PANC) and prostate cancer (PROSTATE) (Fig. \ref{overview}; details in Methods).
We report cMAE as our primary metric because it directly measures the accuracy of individual survival time estimates. We additionally report the c-index to assess risk discrimination, IBS to evaluate time-dependent probabilistic prediction performance, and D-Calibration to assess calibration of the predicted survival distributions.

\subsection*{\ourmethod achieves competitive accuracy for individual survival time estimates}

\begin{figure}[h]
\centering
\includegraphics[width=\textwidth]{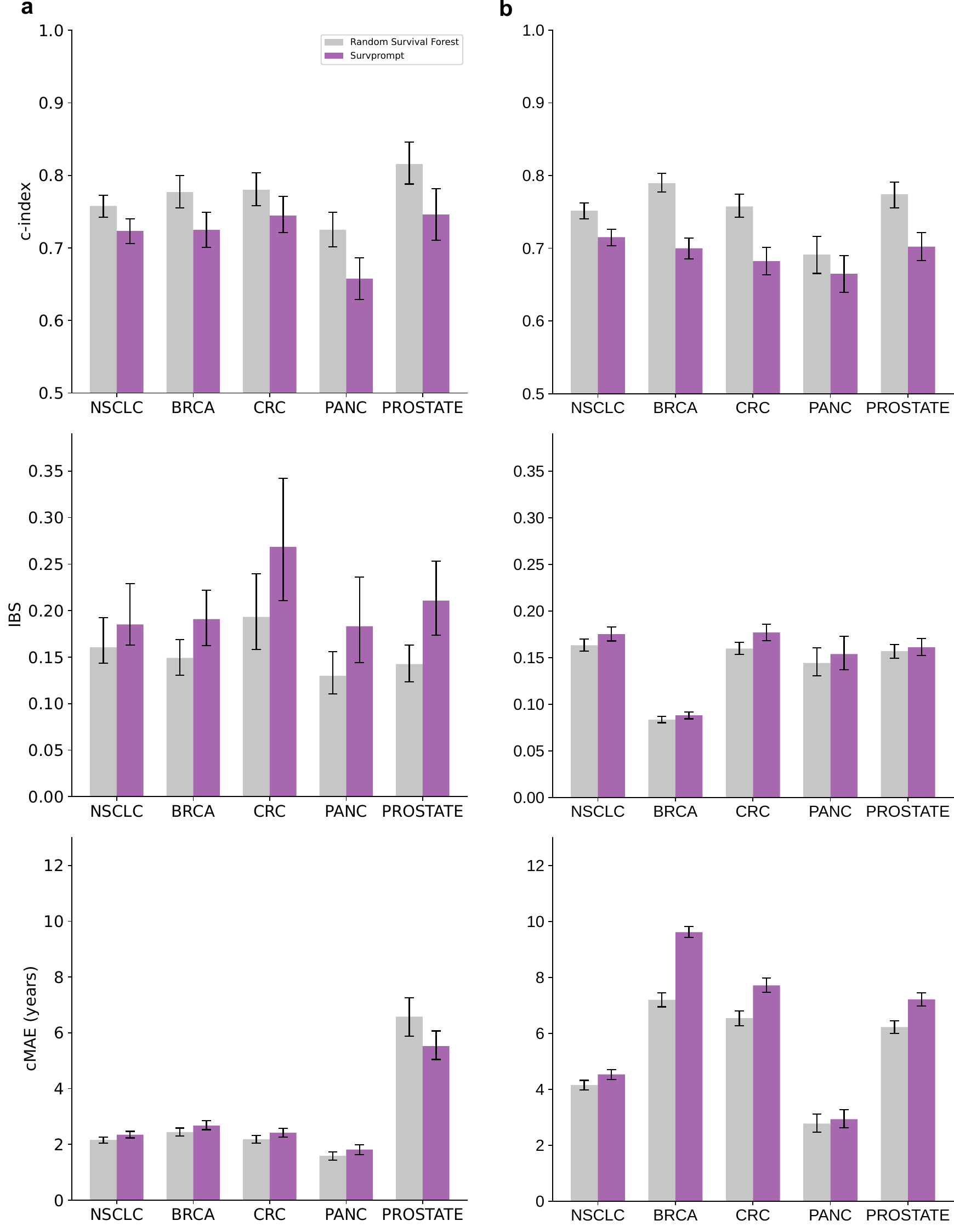}
\caption{\textbf{Summary Performance Metrics across cancer types.} Measures of c-index, integrated Brier score (IBS), and censored mean absolute error (cMAE) comparing the random survival forest baseline and \ourmethod predictions for five cancer types in \textbf{a,} the MSK-CHORD cohort and \textbf{b,} the Providence cohort. Error bars show 95\%  bootstrap confidence intervals for each metric for 1,000 bootstrap samples. NSCLC, non-small cell lung cancer; BRCA, breast cancer; CRC, colorectal cancer; PANC, pancreatic cancer.
}\label{summary_metrics}
\end{figure}
Across both datasets and five cancer types, \ourmethod achieved cMAE values broadly comparable to specialized RSF models trained on structured data (Fig. \ref{summary_metrics}), demonstrating that zero-shot LLMs can generate reasonably accurate individual survival estimates without task-specific training.

Performance varied across cancer types and cohorts.
\ourmethod most closely approached RSF for patients with prostate cancer in MSK-CHORD, where its cMAE point estimate was slightly lower than RSF's (cMAE in years [95\% CI] 5.52 [5.04-6.07] vs 6.58 [5.87-7.25]), and for patients with pancreatic cancer in Providence, where the two were comparable (2.93 [2.63-3.27] vs 2.77 [2.47-3.12]); these indicate that zero-shot LLMs can approach the point-estimate accuracy of fully trained models for some patient groups.
For the remaining cancer types, RSF retained a modest advantage, for example in NSCLC in MSK-CHORD (cMAE in years [95\% CI]: RSF 2.16 [2.05-2.26] vs \ourmethod 2.35 [2.23-2.47]) and colorectal cancer in Providence (RSF 6.54 [6.27-6.79] vs \ourmethod 7.71 [7.47-7.97]).

Despite competitive cMAE, \ourmethod showed a lower c-index than RSF for every cancer type in both cohorts, with the two most comparable for colorectal cancer in MSK-CHORD and pancreatic cancer in Providence (Fig. \ref{summary_metrics}), indicating limited ability to rank patients by risk.
This dissociation was clearest for prostate cancer in MSK-CHORD, where \ourmethod matched RSF on cMAE yet had a substantially lower c-index ([95\% CI] 0.75 [0.70-0.78] vs 0.82 [0.79-0.85]) as well as a worse IBS (Fig. \ref{summary_metrics}), indicating that comparable point-estimate accuracy for this group did not translate to comparable risk discrimination or probabilistic prediction performance.
The IBS followed the same pattern: RSF achieved a lower (better) IBS point estimate than \ourmethod for every cancer type in both cohorts, though the difference was pronounced only in MSK-CHORD and the two were broadly comparable across the Providence cohort (Fig. \ref{summary_metrics}).
Distributional calibration favored RSF (Table \ref{tab:dcalibration}): RSF was well-calibrated across all five cancer types in MSK-CHORD and for pancreatic cancer in Providence, whereas \ourmethod failed the D-Calibration test for every cancer type in both cohorts.
This dissociation reveals a fundamental limitation, namely while \ourmethod can estimate \textit{how long} an individual patient might survive with reasonable accuracy, it struggles to determine who is at higher risk relative to others.

Because API-served LLMs can be nondeterministic, we assessed the stability of \ourmethod by repeating inference three times on MSK-CHORD. Run-to-run variability was small: the mean per-patient standard deviation of predicted survival time (restricted mean over 0 to 10 years) was 0.23 years, and cohort-level metrics varied minimally across runs (per-cancer across-run standard deviations of at most 0.006 for c-index, 0.02 years for cMAE, and 0.001 for IBS).

\subsection*{Qualitative analysis illustrates strengths and weaknesses of \ourmethod}

To assess how well LLM-based predictions align with real-world survival distributions, we compared Kaplan-Meier survival curves generated from model predictions to ground-truth observed outcomes.
Survival models often exhibit over-optimistic or over-pessimistic predictions, reflected in the area under the predicted survival curve being greater than or less than that of the ground-truth curve, respectively \cite{survivalgan}.

\begin{figure}[h]
\centering
\includegraphics[width=\textwidth]{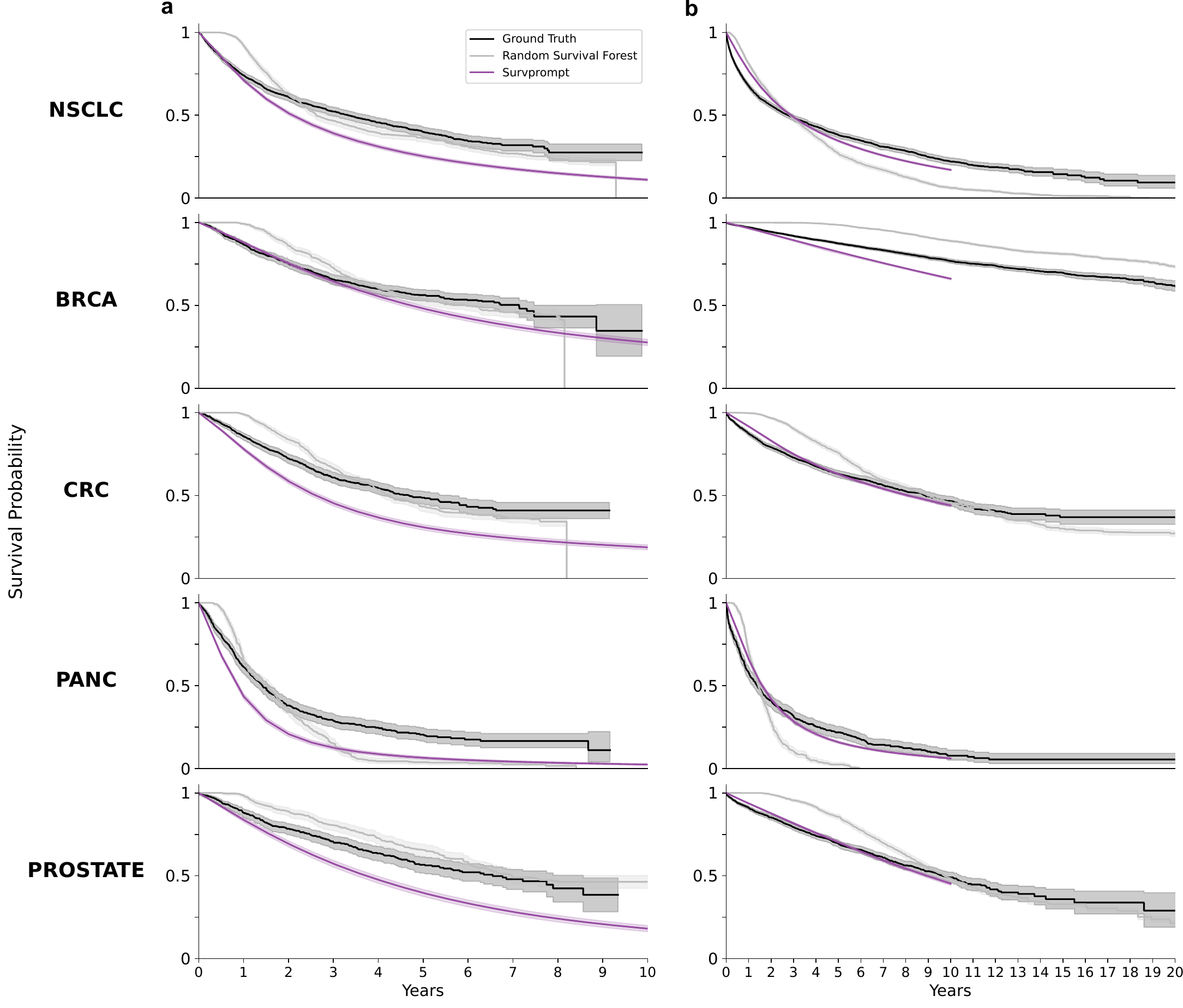}
\caption{\textbf{Performance of \ourmethod in survival prediction across cancer types.} Kaplan-Meier survival curves comparing ground truth observations, random survival forest (RSF) baseline, and \ourmethod predictions for five cancer types in \textbf{a,} the MSK-CHORD cohort and \textbf{b,} the Providence cohort. \ourmethod predictions extend only to the fixed 10-year prompt horizon, whereas the ground-truth and dataset-specific RSF curves span each cohort's full observed follow-up (to 20 years for Providence). NSCLC, non-small cell lung cancer; BRCA, breast cancer; CRC, colorectal cancer; PANC, pancreatic cancer.
}\label{prompt_surv_curves}
\end{figure}

In Figure \ref{prompt_surv_curves}, we compare the ground-truth and RSF-derived Kaplan-Meier curves with the average survival curve predicted by \ourmethod.
Overall, LLM-derived curves approximate the ground truth, though the overlap depends on the cohort, cancer type, and prediction time.
Average survival curves produced by \ourmethod overlap most with the ground truth in early prediction windows ($<$3 years for MSK-CHORD, $<$5 years for Providence) and tend to underestimate survival at longer time horizons.
RSF point estimates initially overestimate survival and, particularly in MSK-CHORD, show increasing fidelity to the ground truth over time, reflecting the advantages of dataset-specific training; in Providence, however, RSF error was less consistent, overestimating survival throughout for breast cancer and drifting toward underestimation at longer horizons for some cancer types.

\begin{figure}[h]
\centering
\includegraphics[width=0.9\textwidth]{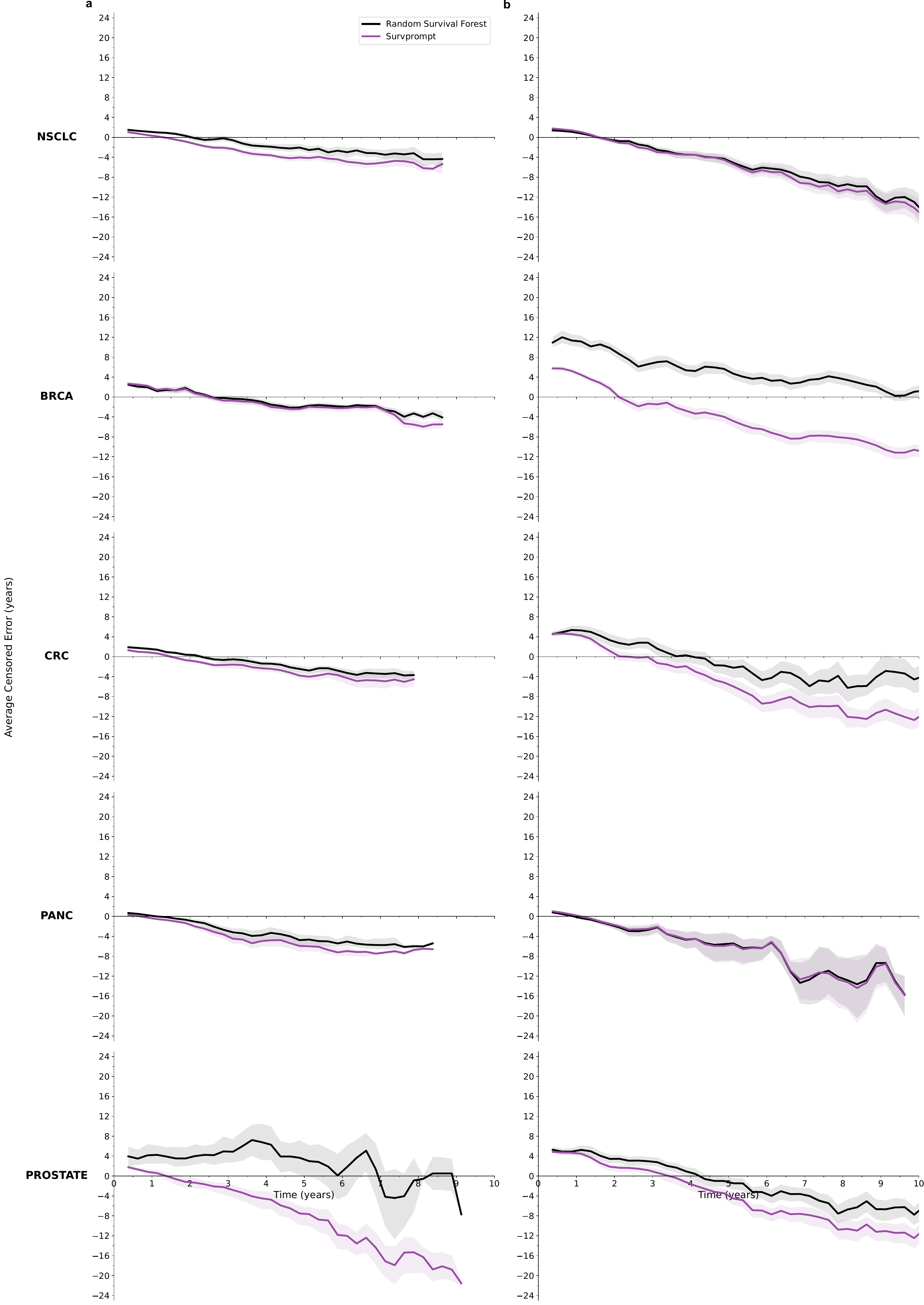}
\caption{\textbf{Time-dependent error visualizations}  Prediction errors of \ourmethod and random survival forest baseline, for five cancer types in \textbf{a,} the MSK-CHORD cohort and \textbf{b,} the Providence cohort. NSCLC, non-small cell lung cancer; BRCA, breast cancer; CRC, colorectal cancer; PANC, pancreatic cancer.
}\label{prompt_surv_errors}
\end{figure}

To further understand time- and dataset-dependent performance, we developed time-aware error visualizations (Figure \ref{prompt_surv_errors}) showing the magnitude and direction of error as a function of prediction horizon.
For both models, errors were smallest at short prediction horizons and grew as the horizon lengthened. \ourmethod increasingly underestimated survival at longer horizons across nearly all cancer types and both cohorts, whereas the direction of RSF error was more cohort- and cancer-dependent.
In several cases, such as patients with breast cancer in Providence and patients with prostate cancer in MSK-CHORD, RSF consistently overestimated survival whereas \ourmethod consistently underestimated it.

\subsection*{\ourmethod exhibits feature reliance patterns similar to traditional survival modeling}

\begin{figure}[h]
\centering
\includegraphics[width=\textwidth]{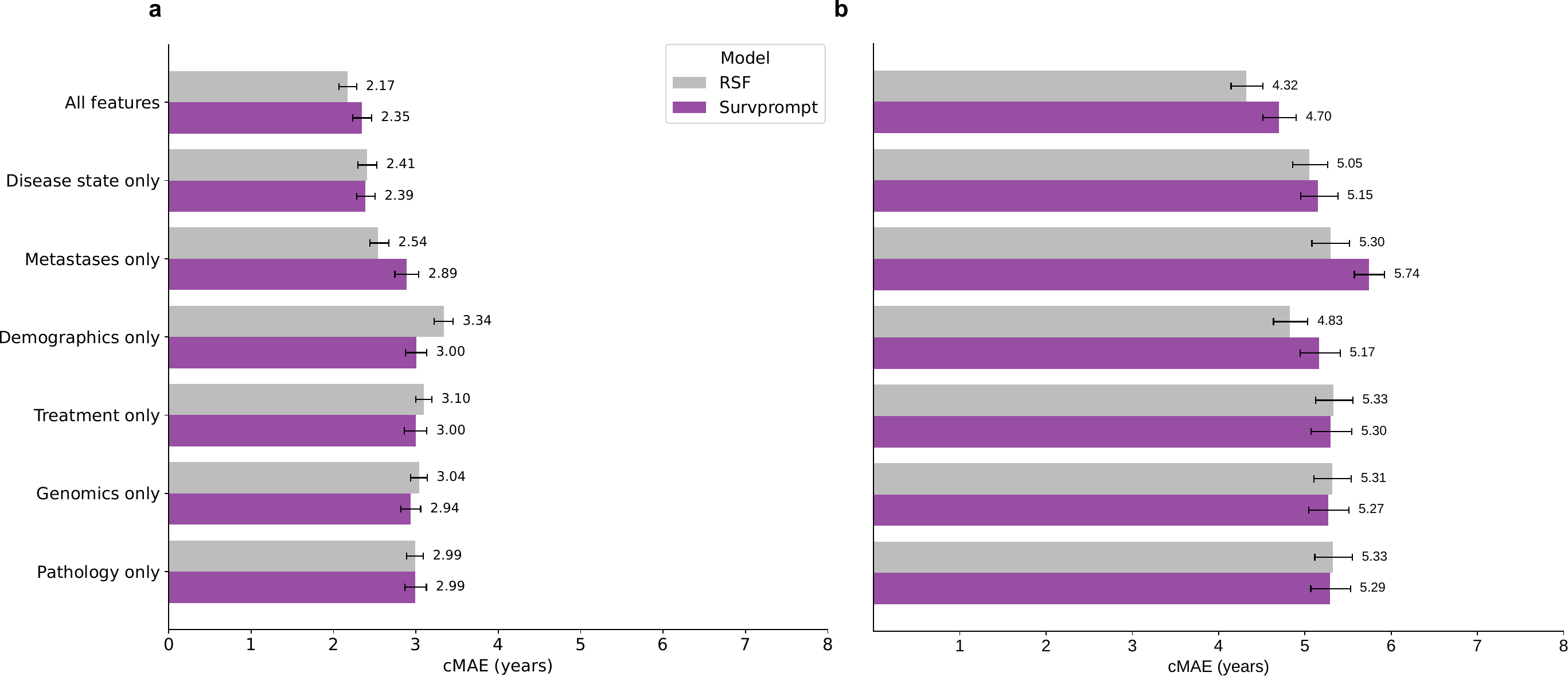}
\caption{\textbf{Ablation study} Comparison of censored MAE (cMAE) values for \ourmethod and RSF for survival estimate prediction task using subsets of features for patients with non-small cell lung cancer in \textbf{a} MSK-CHORD cohort and \textbf{b} Providence cohort.}
\label{ablation_dropout}
\end{figure}

To investigate which variables influence LLM predictions, we evaluated \ourmethod with subsets of features (disease state, metastases, demographics, treatment, genomics, and pathology) and compared with a model with access to all features.
We conducted this ablation in patients with non-small cell lung cancer (NSCLC) across the MSK-CHORD (n=1,562) and Providence (n=3,307) cohorts.

Providing \ourmethod with all features yielded strong performance (cMAE in years [95\% CI] 2.35 [2.23-2.46] in MSK-CHORD, 4.70 [4.51-4.90] in Providence). In MSK-CHORD, the disease-state features alone (e.g., staging and progression) performed comparably (2.39 [2.28-2.50] years), suggesting the LLM draws heavily on disease-state information (Figure \ref{ablation_dropout}).
Single-category subsets performed worse, with treatment-only among the poorest in MSK-CHORD (cMAE in years [95\% CI] 3.00 [2.86-3.13]) and metastases-only the poorest in Providence (5.74 [5.57-5.93]).
RSF showed a similar feature-reliance pattern, performing best with all features (cMAE in years [95\% CI] 2.17 [2.07-2.28] in MSK-CHORD, 4.32 [4.14-4.51] in Providence) and worst with demographics alone in MSK-CHORD (3.34 [3.22-3.45] years) and treatment alone in Providence (5.33 [5.12-5.56]).
A parallel c-index ablation showed a consistent ordering (Table \ref{tab:ablation_cindex}), with disease-state features preserving most of \ourmethod's discrimination in MSK-CHORD (c-index [95\% CI] 0.70 [0.69-0.72] versus 0.72 [0.71-0.74] with all features).
While specialized RSF outperformed \ourmethod in predictive accuracy across most subsets, the alignment in feature importance suggests that LLMs, despite their black-box nature, prioritize clinically relevant information in a manner analogous to specialized survival models.

\section*{Discussion}\label{discussion}
Overall, \ourmethod demonstrates that off-the-shelf LLMs can serve as survival estimators in patients with cancer.
We show that frontier models, in zero-shot form, can provide realistic estimates of survival, while still generally trailing behind task-specific state-of-the-art methods such as RSF in predicting overall survival.
This highlights the ability of pre-trained LLMs to generate accurate survival estimates from free-text descriptions of patient characteristics without task-specific pre-training.
In addition, our results show potential pitfalls when using LLMs to inform patients of their prognosis and thus their output should be interpreted with care.
Our findings provide evidence relevant to clinicians, health systems, AI researchers, regulators, and policymakers as they define safe and effective roles for foundation models in healthcare, while highlighting the need for careful evaluation before clinical deployment.

Prior work has explored the use of LLMs for outcome prediction.
Han et al. \cite{gpt4cardiovascular} compared GPT-4 to existing clinical scores for 10-year cardiovascular risk prediction and found similar discrimination, although others reported poor calibration in the same task \cite{marafinogptcard}.
In oncology, fine-tuned LLMs demonstrated modest c-index improvements over traditional models in a cohort of 1,479 patients \cite{llm4outcomes}, and prior reviews have proposed alternative strategies including LLM-derived embeddings for downstream classifiers \cite{llm4survivalreview}.
We extend these observations by showing that frontier LLMs, without fine-tuning or downstream modeling, can approximate the performance of traditional survival models.

The metrics  we report capture distinct facets of prediction quality, and \ourmethod performed unevenly across them.
For discrimination, the LLM's c-index was consistently below RSF across all cancer types and both cohorts, indicating that although LLMs encode general knowledge about expected survival times, they are less effective at ranking individuals by relative risk.
For point-estimate accuracy, cMAE was broadly comparable to RSF and was lower for breast cancer in the Providence cohort, with non-overlapping confidence intervals.
This point-estimate comparison should nonetheless be interpreted cautiously under heavy censoring. As detailed in Methods, for breast cancer in Providence more than half of patients remained alive at 20 years. Consequently, most pseudo-observation surrogates fall in the poorly supported Kaplan-Meier tail and are capped, compressing the cMAE for that cohort toward a lower bound on the true error.
The IBS, which reflects both discrimination and calibration of the full predicted survival curve, generally favored RSF, consistent with its stronger discrimination, although \ourmethod attained a comparable IBS for prostate cancer in Providence.
Distributional calibration assessed specifically by D-Calibration was cohort-dependent. RSF was well calibrated across MSK-CHORD but not Providence, whereas \ourmethod was well-calibrated for colorectal and prostate cancer in Providence, so neither approach was uniformly better calibrated.
In absolute terms, however, calibration was frequently poor: D-Calibration rejected adequate calibration (p$<$0.05) in the majority of model, cancer type, and cohort combinations, so these well-calibrated cases were the exception rather than the rule, and both models' predicted survival distributions should be treated as imperfectly calibrated.
Overall, this implies that clinically LLMs can approximate broad survival estimates but have limited value for stratifying or prioritizing patients within a cohort where accurate ranking is essential.
Future work could target this ranking weakness directly, for example by fine-tuning LLMs with ranking-aware or concordance-based objectives that better capture relative risk. In data-rich settings where task-specific models can be trained, however, those specialized approaches still dominate, positioning zero-shot LLMs most usefully where such models are unavailable or impractical to build.

Several ethical aspects of survival estimation with \ourmethod warrant discussion.
Our approach offers several potential benefits.
First, it provides individualized, patient-specific survival estimates and is easy for both clinicians and patients to use.
In current clinical practice, clinicians often either avoid giving precise survival prognoses due to the limited accuracy or availability of existing prognostic tools, or rely on broad population-based estimates such as \enquote{$x\%$ of patients with your diagnosis will live up to five years}; an accessible individualized estimator could help address this gap.
Further, frontier LLMs are accessible through application programming interfaces (APIs) or chat interfaces, which may empower patients who want survival information to guide life decisions \cite{patientinfoprefs}.
Second, unlike traditional survival models, our method does not require laborious dataset curation, model training, or maintenance of a collection of bespoke methods.
However, there are also potential risks.
The same direct accessibility that empowers patients may also expose them to psychological harm from inaccurate or poorly contextualized survival estimates \cite{prognosticawareness, llmdelusional}, a risk amplified by the variable accuracy we observe across cancer types and institutions.
In addition, misuse of such predictions could contribute to denial of care by clinicians or insurers \cite{faceage, accurateprediction}.
As with other AI models, equitable deployment of \ourmethod will require careful assessment of potential performance disparities across demographic subgroups, as well as strategies to mitigate bias. In addition, reliable use will depend on the quality and completeness of available clinical information, as inaccurate or missing inputs may compromise survival estimates.
While our study provides strong evidence that LLMs can accurately estimate survival in some settings, further studies are needed before \ourmethod can guide cancer care directly.
Prospective evaluations and assessment of deployment-specific performance will be especially important.

Our work has limitations.
First, interpretability remains limited, and understanding the reasoning behind LLM predictions is an active area of research \cite{biologyllm, llminterpretability}.
While deep learning models are often viewed as less explainable than tree-based methods, there is evidence that the gap in perceived interpretability for end users may be smaller than expected \cite{deeplearnexplainability}.
Second, LLMs served via APIs are subject to nondeterminism in practice \cite{llmnondeterminism2, llmnondeterminism1}; reasoning models such as GPT-5.6-Sol do not expose temperature or other decoding controls, so identical prompts may yield varying outputs across calls.
In practice, however, this variability was small: across three independent runs, both individual predictions and cohort-level metrics were highly stable (Results).
Third, while we demonstrate performance in multiple large, multi-site cancer cohorts, further validation in other populations and disease areas will be important.
Fourth, because MSK-CHORD is publicly available, we cannot exclude the possibility of training-data contamination; however, the comparably strong zero-shot performance on the fully private, held-out Providence cohort, which could not have been part of any pretraining corpus, argues against memorization as the primary explanation for our results.
Fifth, predictors for the Providence cohort were derived from clinician-authored notes using an LLM-based medical abstraction framework \cite{Preston2023,UMA}. While this enables scalable use of routinely collected clinical documentation, abstraction errors may introduce additional uncertainty into downstream survival predictions.
Sixth, although feature-specific ablations assessed the contribution of predictor categories, we did not evaluate disparities in model performance across patient demographic subgroups.
Finally, development of local LLMs with comparable performance could mitigate concerns related to privacy, sustainability, and cost \cite{llavarad}.

\subsection*{Conclusions}
Zero-shot LLMs performed surprisingly well at estimating survival from free-text clinical descriptions of patients with cancer, approximating specialized models on point-estimate accuracy while remaining weaker at discriminating individual risk.
Although dataset- and cancer-specific models such as RSF still outperform LLMs across most discrimination and calibration metrics, the portability of LLMs and their flexibility in handling heterogeneous, free-text inputs make them attractive in data-scarce settings where training a specialized model is impractical.
Given their current limitations and variable accuracy across cancer types and cohorts, however, LLM outputs should be interpreted with care: patients and clinicians consulting them for prognosis should treat the results as rough estimates rather than personalized risk predictions, and further development and prospective validation will be needed before these tools can guide cancer care directly.

\section*{Methods}\label{methods}

\subsection*{Datasets}\label{methods:dataset}

We leverage two real-world cancer datasets to estimate survival.
These datasets span various institutions and states, and are not curated based on specific selection criteria to allow for assessment across larger, general cancer populations.

One dataset is MSK-CHORD \cite{mskchord}, a multi-institutional clinicogenomic dataset containing patients with non-small cell lung (NSCLC; n=7,810), breast (n=5,368), colorectal (n=5,545), pancreatic (n=3,109), or prostate (n=3,211) cancer.
Features cover demographics, staging, sites of metastasis, pathology, genomic mutations across a panel of 23 genes, and prior treatment (full list in Table \ref{tab:feature_extraction}).
Survival data is represented as the number of days from tumor sequencing to death or censoring.

The second dataset is a retrospective cohort assembled from the Providence St. Joseph Health Network covering 51 hospitals and over a thousand clinics across seven U.S. states.
Following institutional review board approval, we assembled this dataset by applying a previously published LLM-based structuring framework \cite{Preston2023,UMA} to patient notes, extracting the same feature set as MSK-CHORD and yielding cohorts of patients with non-small cell lung (n=16,533), breast (n=46,396), colorectal (n=12,793), pancreatic (n=3,539), or prostate (n=11,767) cancer.
Survival data is represented as the number of days from start of the first systemic treatment to death or last contact (censored).

For each cancer type, we split the data into training and evaluation sets (80/20\%). Full feature extraction details and evaluation-set statistics are provided in Supplementary \ref{app:datasets}.

\subsection*{Prompting for Survival Prediction}\label{methods:prompting}
\ourmethod prompts a pre-trained LLM with a free-text clinical vignette constructed from the structured covariates of the MSK-CHORD and Providence datasets (Figure \ref{overview}).
Each prompt consists of system instructions and a vignette describing the patient's presentation.
We instruct the model to produce 21 $(x,y)$ coordinate pairs that parameterize the patient's survival curve, where $x \in \{0, 0.5, 1, \dots, 9.5, 10\}$ years and $y$ is the estimated probability of survival at time $x$.
From these pairs, we extract an expected survival time estimate for each patient.
The prompt horizon is fixed at 0 to 10 years for both cohorts, so \ourmethod emits survival estimates only up to 10 years by design, whereas the dataset- and cancer-specific RSF baseline extends over each cohort's full observed follow-up.
A sample prompt is provided in Figure \ref{fig:sample_prompt_surv_prob}.

We benchmarked several frontier LLMs (GPT-4o \cite{gpt4o}, GPT-4o mini \cite{gpt4omini}, GPT-4.1 \cite{gpt41}, GPT-4.1 mini \cite{gpt41}, o1 \cite{o1}, GPT-5 \cite{gpt5}, GPT-5.4, GPT-5.5, and GPT-5.6-Sol) and selected GPT-5.6-Sol for the main text based on its performance on a held-out validation split and because it returned a valid, appropriately formatted response to every query; all results reported here are computed on a separate test set, and full per-model results are provided in Supplementary \ref{supp_model_comparison}.
All experiments were carried out via private HIPAA-compliant Azure OpenAI endpoints; data shared through these endpoints is not used to train models nor shared externally.

\subsection*{Baseline Approaches}\label{methods:baselines}

We select two strong baselines representing the state-of-the-art methods in survival prediction.
The first baseline consists of random survival forests \cite{rsf}, a decision tree ensemble method designed with custom splitting rules and a conservation-of-events principle for the analysis of right-censored survival data.
This is the model proposed by the original MSK-CHORD publication \cite{mskchord}.
We additionally evaluated a Cox proportional hazards model \cite{coxph}; because RSF was the stronger baseline across cancer types and cohorts, we report RSF as the primary comparator throughout the main text.
Implementation details of baselines are provided in Supplementary \ref{baseline_hyperparams}.

\subsection*{Performance Metrics}
We report a suite of complementary metrics spanning three aspects of survival prediction quality: risk discrimination, point-estimate accuracy, and calibration.

\textit{Discrimination.} The concordance index (c-index) is a rank correlation measure between predicted risk and observed event times: it quantifies whether the model can correctly order patients by relative risk, and is widely used in survival analysis to assess discriminative ability.
However, the c-index does not reflect how accurately a model estimates \textit{when} an event will occur for an individual patient, and has been shown to have important limitations in clinical use \cite{cidxpitfalls}.

\textit{Point-estimate accuracy.} The mean absolute error (MAE) directly measures the magnitude of point-estimate errors in interpretable units of time.
To account for censored observations, we use the pseudo-observations method \cite{pseudoobservations, censoredmae} to compute a surrogate time of death and report a censored MAE (cMAE).
For each censored patient, this imputes an expected event time from the training-set Kaplan-Meier curve; because these surrogates extrapolate the Kaplan-Meier tail, we cap them at the smaller of 30 years and the maximum training follow-up to prevent unsupported extrapolation into regions with no observed data.
This cap materially affects only the Providence breast cancer cohort, where the low event rate places most censored patients' pseudo-values in the tail; the resulting cMAE distribution for that cohort is compressed toward the cap and should be interpreted as a lower bound on the true error rather than a calibrated estimate.
Because our primary use case is individualized prognosis, we use cMAE as our primary metric.

\textit{Calibration.} The integrated Brier score (IBS) is the time-averaged mean squared error between predicted survival probabilities and observed outcomes, capturing both the discrimination and the calibration of the full predicted survival curve \cite{brier}.
To assess distributional calibration, we report D-Calibration \cite{dcalibration}, which tests whether predicted survival probabilities are statistically consistent with observed event times; we report per-cancer, per-cohort p-values, where failing to reject the null (a larger p-value) indicates a well-calibrated model (Table \ref{tab:dcalibration}).

Together, these metrics characterize the model's ability to discriminate between high- and low-risk patients, the accuracy of its individual survival-time estimates, and the calibration of its predicted survival distributions.

\bibliography{bibliography}

\section*{Supplementary Material}\label{supplementary}
\setcounter{table}{0}
\renewcommand*{\tablename}{Supplementary Table}
\renewcommand{\thetable}{S\arabic{table}}
\renewcommand{\theHtable}{supp.\arabic{table}}

\setcounter{figure}{0}
\renewcommand*{\figurename}{Supplementary Figure}
\renewcommand{\thefigure}{S\arabic{figure}}
\renewcommand{\theHfigure}{supp.\arabic{figure}}

\subsection*{Datasets}\label{app:datasets}
The MSK-CHORD dataset was directly obtained from \url{https://github.com/clinical-data-mining/msk-chord-figures-public/}.
Providence is a private dataset, but we explain the filtering and data extraction as such.
We filtered for all patients with a systemic treatment and chose the first entry of their systemic treatment after being diagnosed with a specific cancer type (which we will refer to as the start date).
Labs and ECOG scores listed within the past 6 months of the start date were extracted.
Genomics were binary indicators if each was present within +/- 6 months of the start date.
Any prior treatment includes any mention of chemotherapy, hormone therapy, immune checkpoint inhibitor or systemic treatment before the start date. Given the potential for implausible dates in EHR data and LLM-curated datasets \cite{UMA,Weiskopf2013}, we excluded therapy start dates before the year 2000. For both MSK-CHORD and Providence, we aimed to extract categories and features that are consistent across both datasets. The list of features extracted for both are listed in Table \ref{tab:feature_extraction}.

\begin{table}[htpb!]
\centering
\caption{Extracted features across MSK-CHORD and Providence. In cases where an \textcolor{red}{$\times$} is present, those features were not available in the dataset. Most features are binary indicators, except those with a $\ast$ are categorical and $\dag$ are continuous. CNS stands for central nervous system.}
\label{tab:feature_extraction}
\begin{tabular}{llcc}
\toprule
\textbf{Category} & \textbf{Feature} & \textbf{MSK-CHORD} & \textbf{Providence} \\
\midrule
\multirow{2}{*}{Demographics}
 & Age$^\dag$, Male, Smoker, White, Black, Asian & \checkmark & \checkmark \\
 & Latino, Middle Eastern, Pacific Islander, American Indian & \textcolor{red}{$\times$} & \checkmark \\
\midrule
\multirow{1}{*}{Treatment}
 & Any prior treatment & \checkmark & \checkmark \\
\midrule
\multirow{13}{*}{Pathology}
 & Has Gleason, Gleason$^\ast$ & \checkmark & \textcolor{red}{$\times$} \\
 & Adenocarcinoma, Nonadenocarcinoma  & \checkmark & \textcolor{red}{$\times$} \\
 & Squamous & \checkmark & \textcolor{red}{$\times$} \\
 & Has PDL1 & \checkmark & \checkmark\\
 & PDL1 & \checkmark & \textcolor{red}{$\times$} \\
 & PDL1 low, high, negative & \textcolor{red}{$\times$} & \checkmark \\
 & HR & \checkmark &  \textcolor{red}{$\times$}\\
 & HER2 & \checkmark &  \textcolor{red}{$\times$}\\
 & Rectal & \checkmark &  \textcolor{red}{$\times$}\\
 & Ascending colon & \checkmark & \textcolor{red}{$\times$} \\
 & Cecum & \checkmark &  \textcolor{red}{$\times$}\\
 & Mucinous & \checkmark &  \textcolor{red}{$\times$}\\
 & Has MSI or dMMR & \checkmark & \textcolor{red}{$\times$} \\
\midrule
\multirow{4}{*}{Disease State}
 & Stage 1, 2, 3, 4 & \checkmark & \checkmark \\
 & Stage IV DX, Stage I-III noprog, Stage I-III prog & \checkmark & \textcolor{red}{$\times$} \\
 & Progressed & \checkmark & \textcolor{red}{$\times$} \\
 & ECOG$^\ast$ & \textcolor{red}{$\times$} & \checkmark \\
\midrule
\multirow{9}{*}{Metastases}
 & Adrenal & \checkmark & \textcolor{red}{$\times$} \\
 & Bone & \checkmark & \textcolor{red}{$\times$} \\
 & Brain & \checkmark & \textcolor{red}{$\times$} \\
 & Liver & \checkmark & \textcolor{red}{$\times$} \\
 & Lung & \checkmark & \textcolor{red}{$\times$} \\
 & Lymph & \checkmark & \textcolor{red}{$\times$} \\
 & Other & \checkmark & \textcolor{red}{$\times$} \\
 & Pleura & \checkmark & \textcolor{red}{$\times$} \\
 & CNS & \textcolor{red}{$\times$} & \checkmark \\
\midrule
\multirow{1}{*}{Genomics}
 & \makecell[l]{KRAS, HRAS, RET, MET, GNAQ, PTEN,\\ KIT, EGFR, FGFR1, FGFR2, FGFR3, PDGFRA,\\ ERBB2, TP53, NRAS, NOTCH1, GNA11,\\ CTNNB1, PIK3CA, IDH1, BRAF, ALK, AKT1} & \checkmark & \checkmark \\
\midrule
\multirow{14}{*}{Labs}
 &  Has CA15-3, CA15-3$^\dag$, max CA15-3$^\dag$ & \checkmark & \textcolor{red}{$\times$} \\
 & Has CEA, CEA$^\dag$, max CEA$^\dag$ & \checkmark & \textcolor{red}{$\times$} \\
 & Has PSA, PSA$^\dag$, max PSA$^\dag$ & \checkmark & \textcolor{red}{$\times$} \\
 & Has CA19-9, CA19-9$^\dag$, max CA19-9$^\dag$ & \checkmark & \textcolor{red}{$\times$} \\
 & Bilirubin$^\dag$ & \textcolor{red}{$\times$} & \checkmark \\
 & ALP$^\dag$ & \textcolor{red}{$\times$} & \checkmark \\
 & ALT$^\dag$ & \textcolor{red}{$\times$} & \checkmark \\
 & AST$^\dag$ & \textcolor{red}{$\times$} & \checkmark \\
 & Albumin$^\dag$ & \textcolor{red}{$\times$} & \checkmark \\
 & Hemoglobin$^\dag$ & \textcolor{red}{$\times$} & \checkmark \\
 & Lymphocytes$^\dag$ & \textcolor{red}{$\times$} & \checkmark \\
 & Neutrophils$^\dag$ & \textcolor{red}{$\times$} & \checkmark \\
 & Platelets$^\dag$ & \textcolor{red}{$\times$} & \checkmark \\
 & WBC$^\dag$ & \textcolor{red}{$\times$} & \checkmark \\
\bottomrule
\end{tabular}
\end{table}

We kept the test data for baseline experiments and LLM prompting consistent by using a \textit{KFold} split across $5$ folds, and all test data are $20\%$ of the full dataset described in Section \ref{methods:dataset}.
We provide an overview of the test data for LLM prompting for MSK-CHORD in Table \ref{tab:mskchord_stats} and for Providence in Table \ref{tab:prov_stats}.
Missingness is present in the Providence dataset, so the percentages refer to the percent of non-missing entries for that feature across cancer type.
Mean values represent the mean where missing entries are dropped.
Progression is not available in Providence and ECOG scores are not available in MSK-CHORD, hence the differing features.

\begin{table}[htpb!]
\caption{Test data statistics by cancer type for MSK-CHORD. SD stands for standard deviation, met stands for metastasis. NSCLC, non-small cell lung cancer; BRCA, breast cancer; CRC, colorectal cancer; PANC, pancreatic cancer.}
\label{tab:mskchord_stats}
\begin{tabular}{llllll}
\toprule
Cancer type & NSCLC & BRCA  & CRC & PANC & PROSTATE \\
\midrule
n & 1562 & 1074 & 1109 & 622 & 643 \\
Progressed & 56.2\% & 50.1\% & 44.6\% & 48.4\% & 60.8\%\\
Died & 51.2\% & 38.4\% & 38.2\% & 66.9\% & 34.7\%\\
Mean Survival in yrs (SD) & 2.5 (2.1) & 3.2 (2.2) & 2.6 (2.0) & 1.7 (1.7) & 3.1 (2.2)\\
Mean Age (SD) & 67 (11) & 56 (13) & 58 (15) & 66 (11) & 67 (9)\\
Male & 41.2\% & 1.1\% & 53.9\% & 52.1\% & 99.8\% \\
White & 79.4\% & 72.9\% & 77.4\% & 81.8\% & 80.2\% \\
Stage 1 & 33.1\% & 33.1\% & 6.9\% & 20.3\% & 7.2\% \\
Stage 2 & 6.9\% & 28.1\% & 13.3\% & 22.2\% & 37.0\% \\
Stage 3 & 16.2\% & 15.5\% & 36.0\% & 15.0\% & 20.5\% \\
Stage 4 & 43.0\% & 20.5\% & 42.8\% & 42.6\% & 35.3\% \\
Met. to adrenal & 10.8\% & 2.8\% & 2.1\% & 4.3\% & 2.6\% \\
Met. to bone & 31.0\% & 43.6\% & 7.6\% & 6.8\% & 51.5\% \\
Met. to brain & 20.4\% & 13.3\% & 3.2\% & 0.5\% & 7.6\% \\
Met. to liver & 14.3\% & 23.9\% & 46.0\% & 42.1\% & 7.5\% \\
Met. to lung & 82.7\% & 24.8\% & 27.9\% & 18.5\% & 24.4\% \\
Met. to lymph & 53.9\% & 47.2\% & 47.3\% & 24.1\% & 52.6\% \\
Met. to other & 36.6\% & 66.1\% & 74.0\% & 83.4\% & 69.5\% \\
Met. to pleura & 17.0\% & 8.2\% & 1.4\% & 1.6\% & 2.0\% \\
\bottomrule
\end{tabular}
\end{table}

\begin{table}[htpb!]
\caption{Test data statistics by cancer type for Providence. SD stands for standard deviation, met stands for metastasis. and CNS stands for central nervous system. NSCLC, non-small cell lung cancer; BRCA, breast cancer; CRC, colorectal cancer; PANC, pancreatic cancer.}
\label{tab:prov_stats}
\begin{tabular}{llllll}
\toprule
Cancer type & NSCLC & BRCA & CRC & PANC & PROSTATE \\
\midrule
n & 3307 & 9280 & 2559 & 708 & 2354 \\
ECOG (SD) & 1.1 (0.9) & 0.4 (0.7) & 0.7 (0.8) & 1.1 (1.0) & 0.9 (1.0) \\
Died & 62.2\% & 16.2\% & 37.5\% & 73.9\% & 33.3\% \\
Mean Survival in yrs (SD) & 3.0 (3.4) & 6.2 (4.7) & 4.3 (3.8) & 2.1 (2.7) & 4.5 (3.4) \\
Mean Age (SD) & 69 (10) & 62 (13) & 64 (14) & 67 (11) & 72 (9) \\
Male & 45.5\% & 0.5\% & 49.5\% & 51.3\% & 99.9\% \\
White & 46.3\% & 40.3\% & 40.1\% & 45.6\% & 47.6\% \\
Stage 1 & 12.9\% & 29.0\% & 9.3\% & 6.4\% & 7.1\% \\
Stage 2 & 4.9\% & 16.5\% & 13.6\% & 12.9\% & 36.5\% \\
Stage 3 & 10.3\% & 5.0\% & 17.6\% & 4.9\% & 10.6\% \\
Stage 4 & 22.5\% & 2.0\% & 9.9\% & 17.8\% & 10.4\% \\
Met. to CNS & 7.7\% & 0.4\% & 0.4\% & 0.6\% & 0.2\% \\
\bottomrule
\end{tabular}
\end{table}

\subsection*{Prompt Generation}
The patient vignette creation pipeline converts structured data into a paragraph format, which is easier to input into a LLM.
This pipeline consists of converting each of the $7$ categories, Demographics, Treatment, Pathology, Disease State, Metastases, Genomics, and Labs, into separate sentences. Missing feature values are omitted from the vignette rather than being reported as unknown.
For example in the Demographics category, a data entry containing $Age=67$, $Male=0$, $Asian=1$ and $Smoker=1$, would be converted into \enquote{The patient is a 67 years old Asian female with a history of smoking}.
Figure \ref{overview} shows an example construction of the full free-text vignette from a sample structured data entry.

Regarding prompting tasks, we provide sample prompts used for the MSK-CHORD dataset using the two different prompting strategies: time-to-event (TTE; Figure \ref{fig:sample_prompt_tte}) and survival-probability (SURV\_PROB; Figure \ref{fig:sample_prompt_surv_prob}).

\begin{figure}[htbp!]
\centering
\begin{tcolorbox}[
    colback=gray!10,
    colframe=gray!50,
    title={\textbf{TTE Sample Prompt}},
    fonttitle=\bfseries,
    boxrule=1pt,
    rounded corners,
]

\textbf{Instructions:} You are an expert oncologist with deep knowledge of treatment modalities, prognostic models and survival statistics.

You will be presented with a clinical vignette describing a patient with cancer, and your task is to estimate the patient's survival time in days. This is the median survival time for the given patient characteristics. Accuracy is the ONLY important factor. Carefully review the vignette, considering all relevant prognostic factors. The provided information may be incomplete, but make the most reasonable assumptions possible.

After \enquote{Estimated survival time (days)}:, you must provide a single integer representing the most accurate estimate.

\textbf{Vignette:}
The patient is a 58 years old, white female with a history of smoking. The patient has been
diagnosed with stage 3 non-squamous cell, adenocarcinoma non-small cell lung cancer (NSCLC). The patient has received prior treatment. A recent radiology report shows: Stage I-III disease with evidence of progression with evidence of metastasis to: lymph nodes, other sites. Pathology report shows: PDL1 expression not assessed. Genomic testing shows oncogenic alterations (mutations, copy number changes and/or structural variations) in the following genes: TP53. There are no oncogenic alterations in the following genes: KRAS, HRAS, RET, MET, GNAQ, PTEN, KIT, EGFR, FGFR1, FGFR2, FGFR3, PDGFRA, ERBB2, NRAS, NOTCH1, GNA11, CTNNB1, PIK3CA, IDH1, BRAF, ALK, AKT1.
\end{tcolorbox}
\caption{}
\label{fig:sample_prompt_tte}
\end{figure}

\begin{figure}[htbp!]
\centering
\begin{tcolorbox}[
    colback=gray!10,
    colframe=gray!50,
    title={\textbf{SURV\_PROB Sample Prompt}},
    fonttitle=\bfseries,
    boxrule=1pt,
    rounded corners,
]

\textbf{Instructions:} You are an expert oncologist with deep knowledge of treatment modalities, prognostic models and survival statistics.

You will be presented with a clinical vignette describing a patient with cancer, and your task is to estimate the patient\'s survival probability for 0 to 10 years in the future. Accuracy is the ONLY important factor. Note that this patient may live less or more than 10 years, but you are required to provide estimates for the first 10 years. This is the probability (decimal ranging from 0 to 1) you\'d expect the patient to survive at each future point in time.

Carefully review the clinical vignette, considering all relevant prognostic factors. The provided information may be incomplete, but make the most reasonable assumptions possible.

Explain your thought process carefully including any assumptions. Then, after \enquote{Survival prediction:}, you must generate 21 (x,y) coordinates. x is the time in years from now and y is your estimated probability of survival (a float value between 0.0 and 1.0) at the corresponding year x. Follow the format: \enquote{$<Explanation>$. Survival prediction: [(0.0, 1.0), (0.5, estimate), (1.0, estimate), (1.5, estimate), (2.0, estimate), (2.5, estimate), (3.0, estimate), (3.5, estimate), (4.0, estimate), (4.5, estimate), (5.0, estimate), (5.5, estimate), (6.0, estimate), (6.5, estimate), (7.0, estimate), (7.5, estimate), (8.0, estimate), (8.5, estimate), (9.0, estimate), (9.5, estimate), (10.0, estimate)]}

\textbf{Vignette:}
The patient is a 58 years old, white female with a history of smoking. The patient has been
diagnosed with stage 3 non-squamous cell, adenocarcinoma non-small cell lung cancer (NSCLC). The patient has received prior treatment. A recent radiology report shows: Stage I-III disease with evidence of progression with evidence of metastasis to: lymph nodes, other sites. Pathology report shows: PDL1 expression not assessed. Genomic testing shows oncogenic alterations (mutations, copy number changes and/or structural variations) in the following genes: TP53. There are no oncogenic alterations in the following genes: KRAS, HRAS, RET, MET, GNAQ, PTEN, KIT, EGFR, FGFR1, FGFR2, FGFR3, PDGFRA, ERBB2, NRAS, NOTCH1, GNA11, CTNNB1, PIK3CA, IDH1, BRAF, ALK, AKT1.

\end{tcolorbox}
\caption{}
\label{fig:sample_prompt_surv_prob}
\end{figure}

\subsection*{Experiments}
\subsubsection*{Baselines}
\label{baseline_hyperparams}
Baseline models were trained using Python 3.10.
Random survival forests were trained using 1,000 trees, requiring at least 10 samples to split an internal node and 15 samples per terminal node.
Trees were grown without a maximum depth constraint, using bootstrap sampling and the log-rank splitting rule, with the number of features considered at each split set to the square root of the total features.
All computations were parallelized across available processors, with a fixed random seed for reproducibility, using the scikit-survival version 0.20.0 \cite{sksurv}.
Cox proportional hazards models were fit with the baseline hazard estimated non-parametrically using Breslow’s method and an $L_2$ penalty of 0.1, without stratification using the lifelines package version 0.27.7 \cite{lifelines}. Missing feature values were set to zero, effectively encoding missing values as the absent/negative class; Cox models additionally excluded samples with missing or non-positive survival duration or event labels. This follows the reference random survival forest implementation proposed by the MSK-CHORD authors, where continuous features with potential missingness (PD-L1 and tumor-marker labs) were each paired with a binary indicator marking whether the value was observed \cite{mskchord}.

\subsubsection*{LLMs}
In addition to the prompting strategy described in the main text (i.e. SURV-PROB), where the model is prompted to predict survival probabilities for a set of pre-determined time points, we explored directly asking the model to provide an expected survival in number of days (TTE). LLM experiments were carried out in HIPAA compliant Azure OpenAI instances using API version.
We provide a table of the model configurations we used for all our experiments and datasets (Table \ref{tab:model_configs}).
\begin{table}[htbp!]
\centering
\caption{Zero-Shot Model Configuration Summary. Reasoning models were queried at their default reasoning effort, for which sampling temperature is not applicable (N/A).}
\label{tab:model_configs}
\begin{tabular}{lccc}
\toprule
\textbf{Model} & \textbf{Version} & \textbf{Temperature} & \textbf{Reasoning} \\ \hline
GPT-4o & 2024-08-06 & 0 & N/A \\
GPT-4o-mini & 2024-07-18 & 0 & N/A \\
GPT-4.1 & 2025-04-14 & 0 & N/A \\
GPT-4.1-mini & 2025-04-14 & 0 & N/A \\
GPT-5 & 2025-08-07 & N/A & Medium \\
GPT-5.4 & 2026-03-05 & N/A & Medium \\
GPT-5.5 & 2026-04-24 & N/A & Medium \\
GPT-5.6-Sol & 2026-07-09 & N/A & Medium \\
o1 & 2024-12-17 & N/A & Medium \\
\bottomrule
\end{tabular}
\end{table}

\section*{Results}
\subsection*{Performance across other LLMs}
\label{supp_model_comparison}

\begin{figure}[h]
\centering
\includegraphics[width=\textwidth]{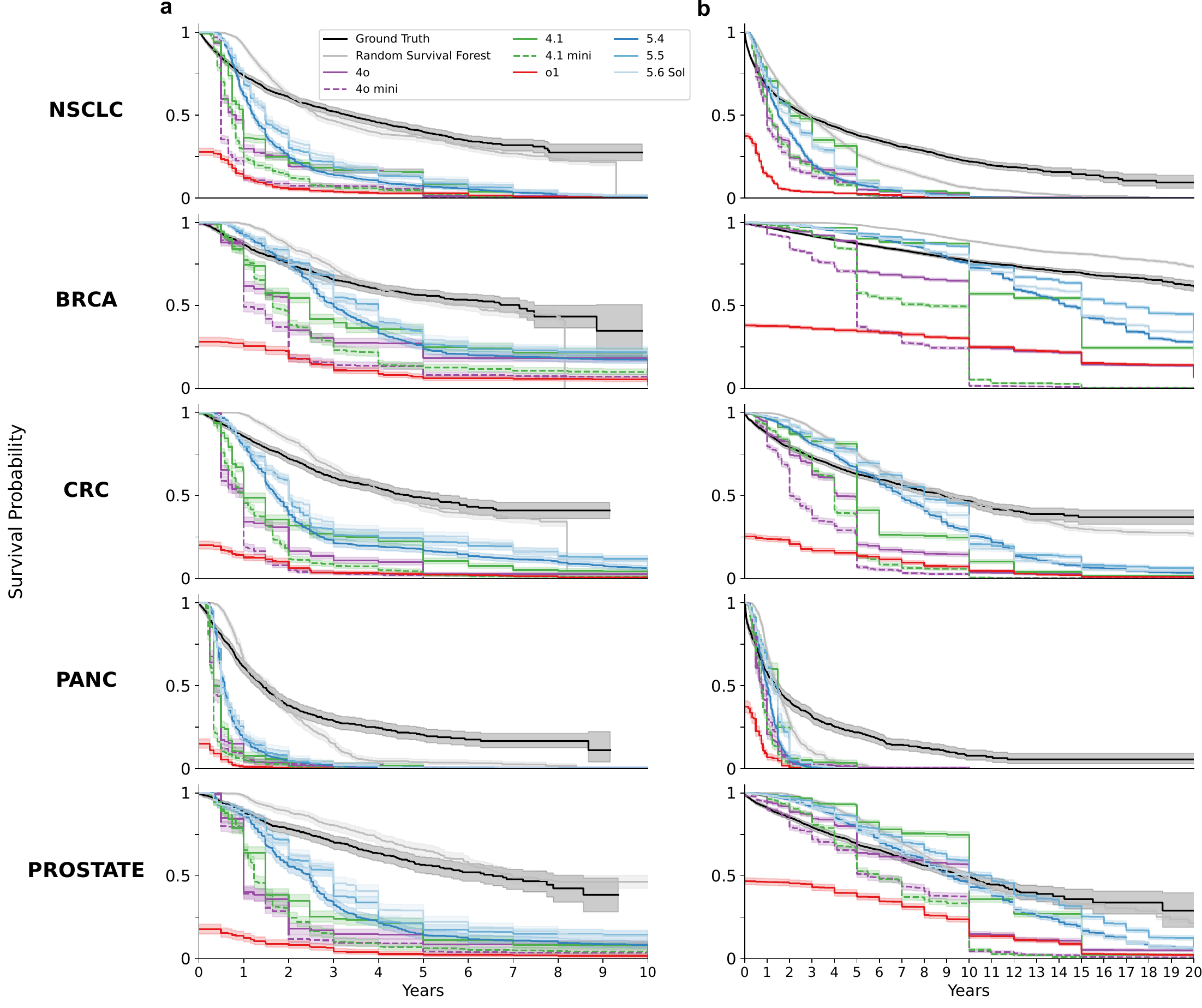}
\caption{\textbf{Performance of \ourmethod-TTE in survival prediction across cancer types.} Kaplan-Meier survival curves comparing ground truth observations, random survival forest (RSF) baseline, and predictions of \ourmethod-TTE for various models for five cancer types in \textbf{a,} the MSK-CHORD cohort and \textbf{b,} the Providence cohort. NSCLC, non-small cell lung cancer; BRCA, breast cancer; CRC, colorectal cancer; PANC, pancreatic cancer.}\label{km_curve_all_tte}
\end{figure}

\begin{figure}[htpb!]
\centering
\includegraphics[width=\textwidth]{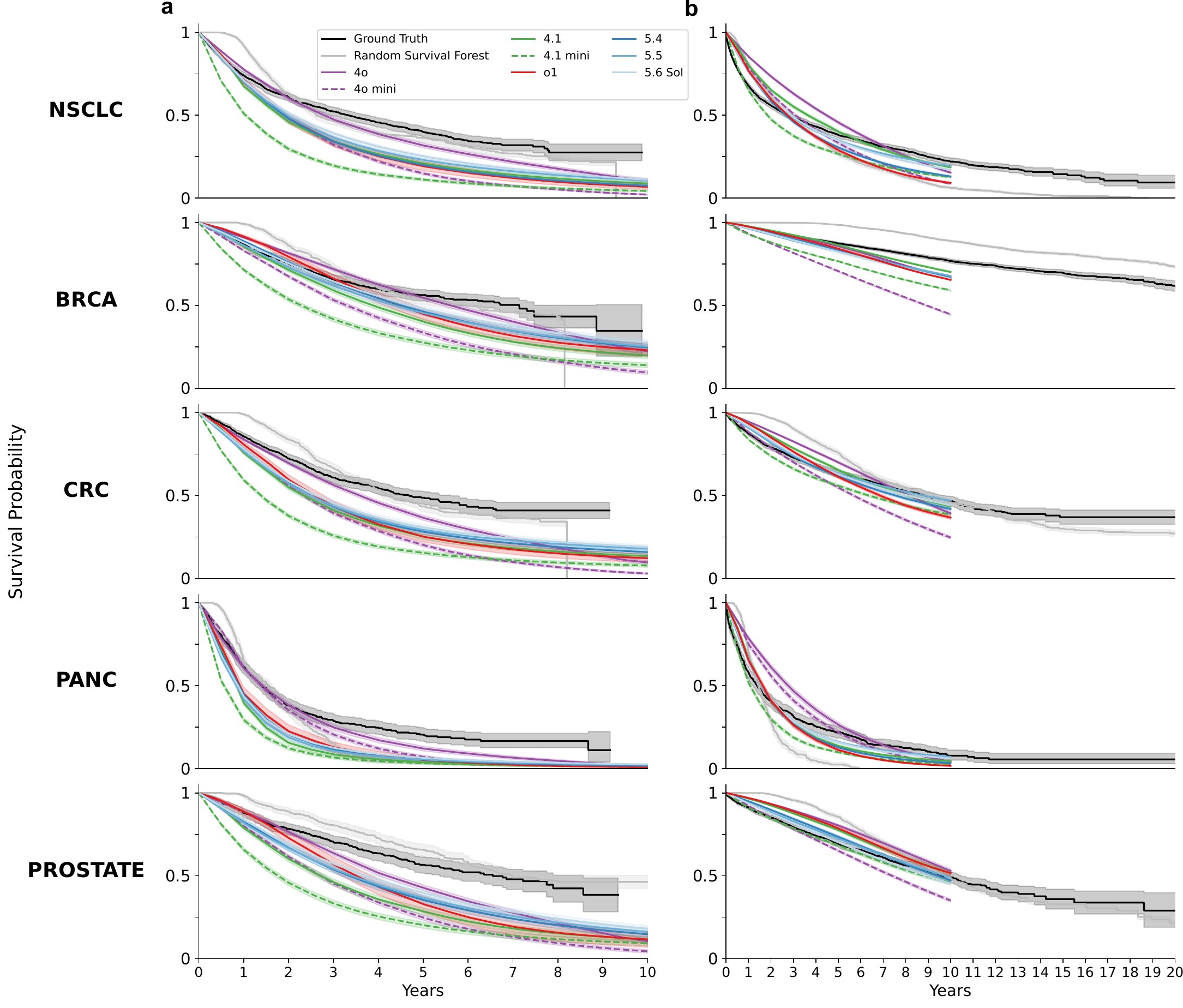}
\caption{\textbf{Performance of \ourmethod-SURV\_PROB in survival prediction across cancer types.} Kaplan-Meier survival curves comparing ground truth observations, random survival forest (RSF) baseline, and predictions of \ourmethod-SURV\_PROB for various models for five cancer types in \textbf{a,} the MSK-CHORD cohort and \textbf{b,} the Providence cohort. NSCLC, non-small cell lung cancer; BRCA, breast cancer; CRC, colorectal cancer; PANC, pancreatic cancer.}\label{km_curve_all_sprob}
\end{figure}

\begin{figure}[htpb!]
\centering
\includegraphics[width=0.9\textwidth]{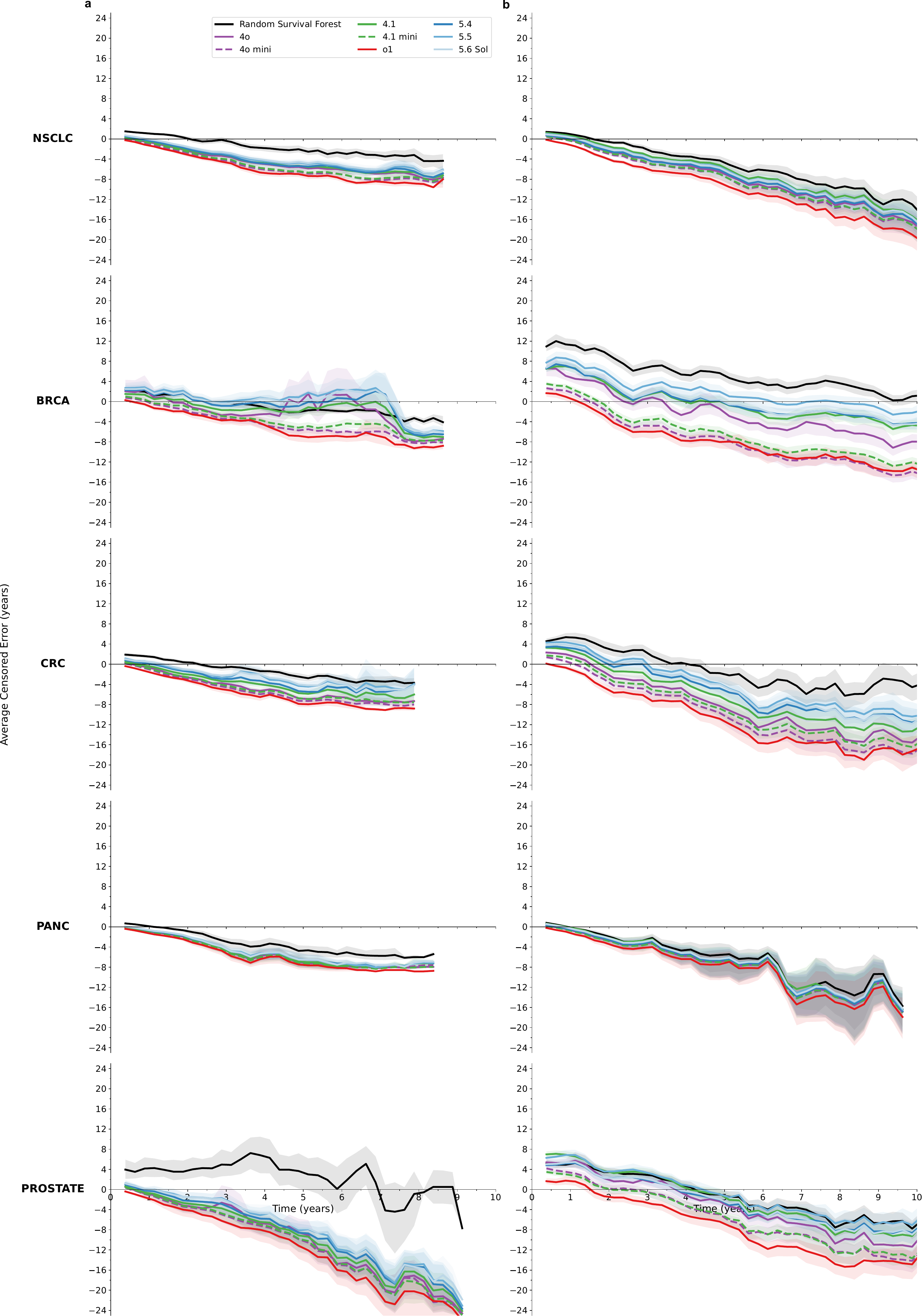}
\caption{\textbf{Time-dependent error visualizations}  Prediction errors of \ourmethod-TTE and random survival forest baseline, for various models for five cancer types in \textbf{a}, the MSK-CHORD cohort and \textbf{b}, the Providence cohort. NSCLC, non-small cell lung cancer; BRCA, breast cancer; CRC, colorectal cancer; PANC, pancreatic cancer.}\label{errors_all_tte}
\end{figure}

\begin{figure}[htpb!]
\centering
\includegraphics[width=0.9\textwidth]{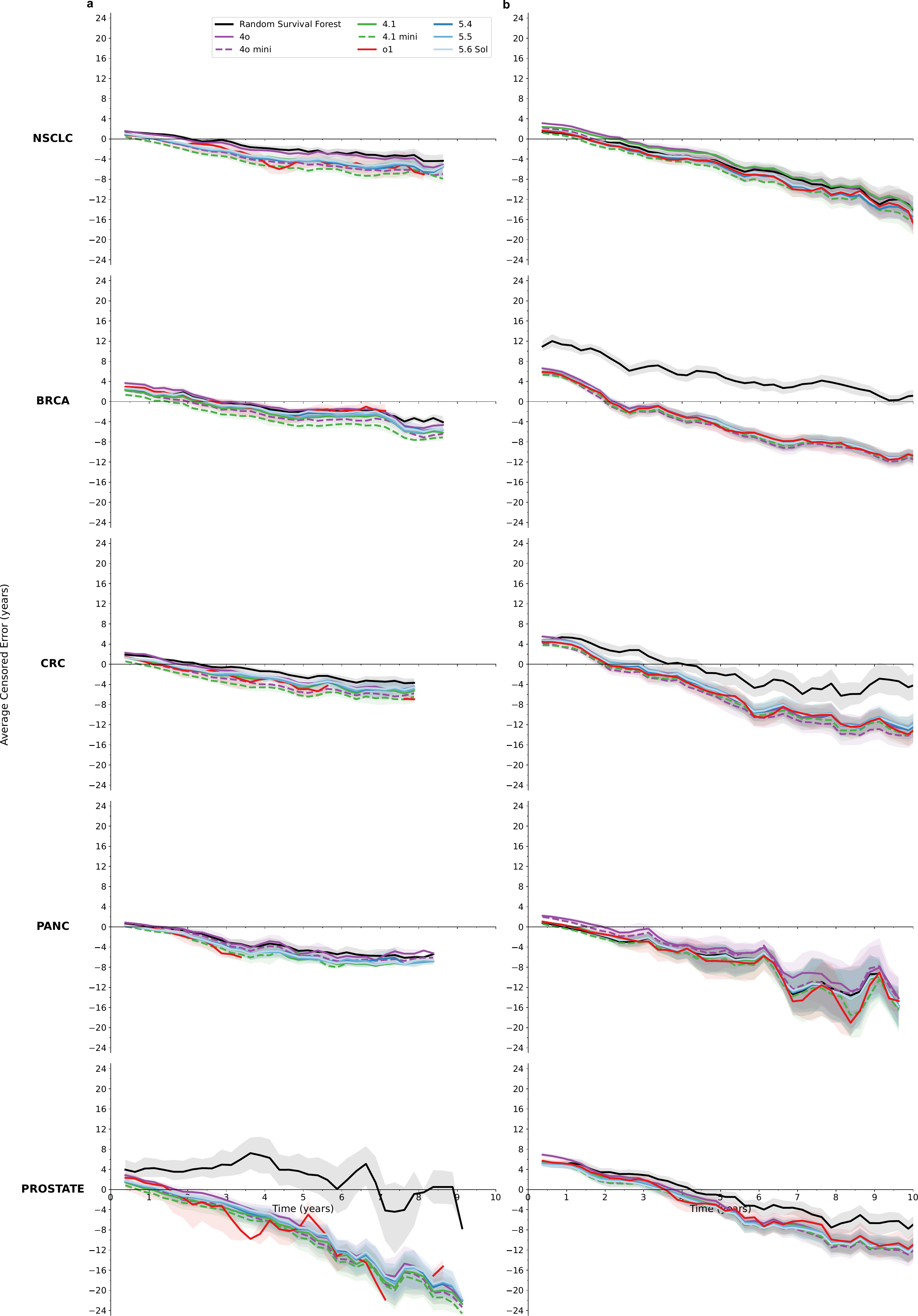}
\caption{\textbf{Time-dependent error visualizations}  Prediction errors of \ourmethod-SURV\_PROB and random survival forest baseline, for various models for five cancer types in \textbf{a}, the MSK-CHORD cohort and \textbf{b}, the Providence cohort. NSCLC, non-small cell lung cancer; BRCA, breast cancer; CRC, colorectal cancer; PANC, pancreatic cancer.}\label{errors_all_sprob}
\end{figure}

\begin{figure}[htpb!]
\centering
\includegraphics[width=\textwidth]{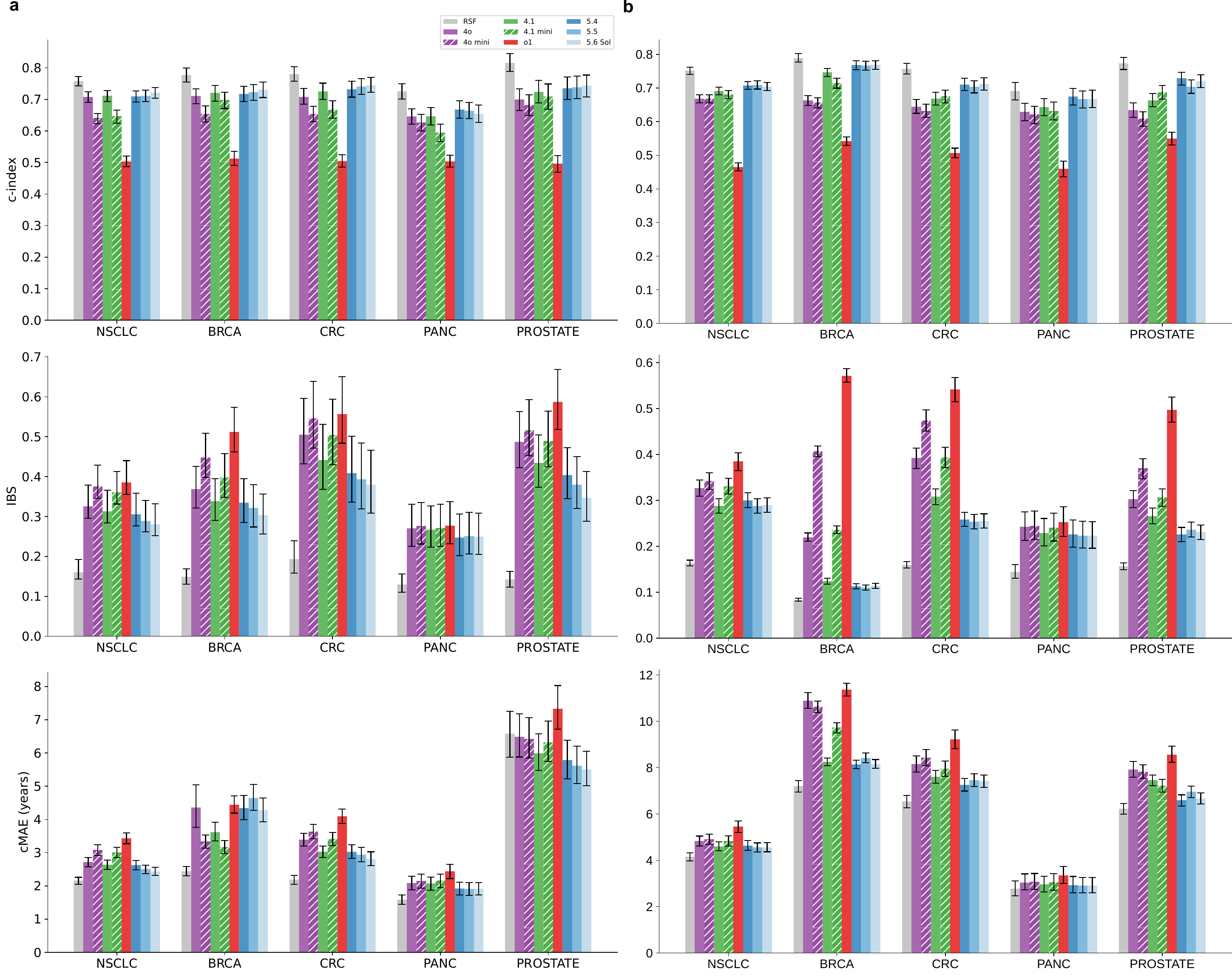}
\caption{\textbf{Summary Performance Metrics across cancer types for \ourmethod-TTE} Measures of c-index, integrated Brier score (IBS), and censored mean absolute error (cMAE) comparing the random survival forest (RSF) baseline, and \ourmethod-TTE predictions for various models for five cancer types in \textbf{a}, the MSK-CHORD cohort and \textbf{b}, the Providence cohort. Error bars show 95\% bootstrap confidence intervals for each metric for 1,000 bootstrap samples. NSCLC, non-small cell lung cancer; BRCA, breast cancer; CRC, colorectal cancer; PANC, pancreatic cancer.}\label{metrics_tte}
\end{figure}

\begin{figure}[htbp!]
\centering
\includegraphics[width=\textwidth]{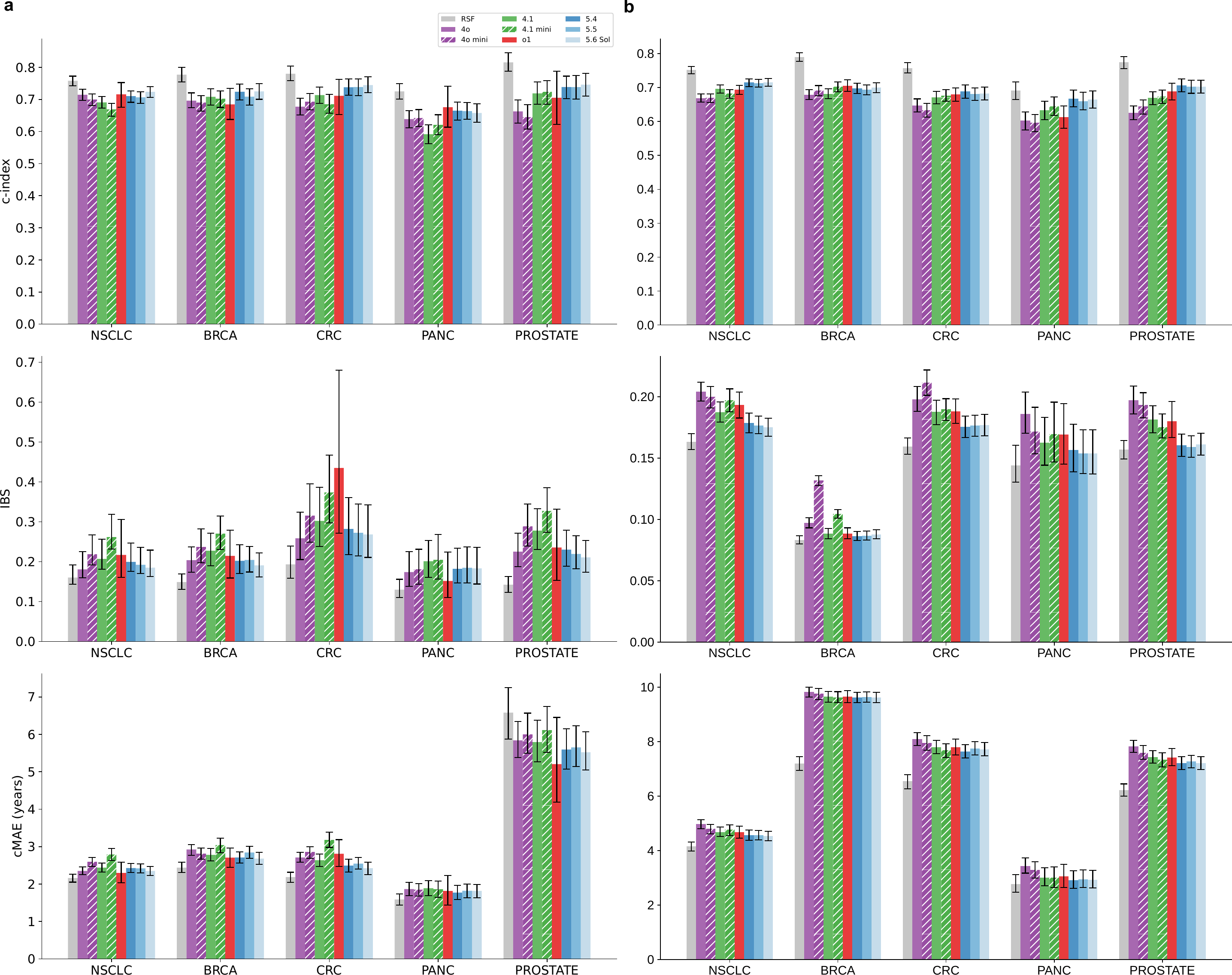}
\caption{\textbf{Summary Performance Metrics across cancer types for \ourmethod-SURV\_PROB} Measures of c-index, integrated Brier score (IBS), and censored mean absolute error (cMAE) comparing the random survival forest (RSF) baseline, and \ourmethod-SURV\_PROB predictions for various models for five cancer types in \textbf{a}, the MSK-CHORD cohort and \textbf{b}, the Providence cohort. Error bars show 95\% bootstrap confidence intervals for each metric for 1,000 bootstrap samples. NSCLC, non-small cell lung cancer; BRCA, breast cancer; CRC, colorectal cancer; PANC, pancreatic cancer.}\label{metrics_sprob}
\end{figure}

We repeated the evaluation across all benchmarked LLMs (Table \ref{tab:model_configs}) under both prompting strategies.
Figures \ref{km_curve_all_tte} and \ref{km_curve_all_sprob} show Kaplan-Meier survival curves for the TTE and SURV\_PROB strategies, respectively; Figures \ref{errors_all_tte} and \ref{errors_all_sprob} show the corresponding time-dependent error visualizations; and Figures \ref{metrics_tte} and \ref{metrics_sprob} summarize per-model performance across cancer types and cohorts.

\begin{table}[htbp!]
\centering
\caption{\textbf{D-Calibration p-values by cancer type and cohort.} P-values from the D-Calibration test \cite{dcalibration} assessing the distributional calibration of \ourmethod and the RSF baseline. Larger p-values (failure to reject the null) indicate better-calibrated predicted survival distributions. NSCLC, non-small cell lung cancer; BRCA, breast cancer; CRC, colorectal cancer; PANC, pancreatic cancer.}
\label{tab:dcalibration}
\begin{tabular}{lcccc}
\toprule
 & \multicolumn{2}{c}{\textbf{MSK-CHORD}} & \multicolumn{2}{c}{\textbf{Providence}} \\
\cmidrule(lr){2-3} \cmidrule(lr){4-5}
\textbf{Cancer} & RSF & \ourmethod & RSF & \ourmethod \\
\midrule
NSCLC    & 0.33 & $<$0.001 & $<$0.001 & $<$0.001 \\
BRCA     & 0.13 & 0.01     & $<$0.001 & $<$0.001 \\
CRC      & 0.27 & $<$0.001 & $<$0.001 & $<$0.001 \\
PANC     & 0.40 & $<$0.001 & 0.78     & $<$0.001 \\
PROSTATE & 0.08 & $<$0.001 & $<$0.001 & $<$0.001 \\
\bottomrule
\end{tabular}
\end{table}

\begin{table}[htbp!]
\centering
\small
\caption{\textbf{Feature-subset ablation: c-index by cohort and model.} Concordance index (c-index) for \ourmethod and the random survival forest (RSF) baseline when restricted to each feature subset, for patients with non-small cell lung cancer (NSCLC) in the MSK-CHORD and Providence cohorts. Values are point estimates with 95\% bootstrap confidence intervals in brackets. This c-index ablation complements the cMAE ablation in Figure \ref{ablation_dropout}.}
\label{tab:ablation_cindex}
\begin{tabular}{lcccc}
\toprule
 & \multicolumn{2}{c}{\textbf{MSK-CHORD}} & \multicolumn{2}{c}{\textbf{Providence}} \\
\cmidrule(lr){2-3} \cmidrule(lr){4-5}
\textbf{Feature subset} & RSF & \ourmethod & RSF & \ourmethod \\
\midrule
All features       & 0.76 [0.74, 0.77] & 0.72 [0.71, 0.74] & 0.75 [0.74, 0.76] & 0.71 [0.70, 0.73] \\
Disease state only & 0.70 [0.69, 0.72] & 0.70 [0.69, 0.72] & 0.64 [0.63, 0.65] & 0.62 [0.61, 0.63] \\
Metastases only    & 0.70 [0.68, 0.72] & 0.68 [0.66, 0.70] & 0.53 [0.52, 0.53] & 0.51 [0.50, 0.53] \\
Demographics only  & 0.53 [0.51, 0.55] & 0.52 [0.50, 0.54] & 0.67 [0.66, 0.68] & 0.60 [0.59, 0.61] \\
Treatment only     & 0.55 [0.54, 0.57] & 0.56 [0.54, 0.58] & 0.50 [0.49, 0.52] & 0.51 [0.49, 0.52] \\
Genomics only      & 0.58 [0.56, 0.60] & 0.56 [0.54, 0.58] & 0.51 [0.51, 0.52] & 0.51 [0.50, 0.52] \\
Pathology only     & 0.58 [0.56, 0.60] & 0.47 [0.45, 0.49] & 0.51 [0.50, 0.52] & 0.50 [0.49, 0.51] \\
\bottomrule
\end{tabular}
\end{table}

\newpage
\section*{Data Availability}\label{data_availability}

The MSK-CHORD dataset is publicly available at \url{https://datacatalog.mskcc.org/dataset/11458}. The Providence dataset is not publicly available due to privacy and compliance considerations established by the research protocol. Queries about either dataset can be directed to the corresponding authors indicated above.

\section*{Code Availability}\label{code_availability}

\ourmethod, the library developed for this project, will be made publicly available at \url{https://aka.ms/survprompt}

\section*{Acknowledgements}\label{acknowledgements}

The authors thank the patients who contributed data to this study.

\section*{Funding}\label{funding}
This work was supported by Microsoft Research.

\section*{Author Contributions}\label{author_contributions}
J.M.Z.C., P.A.\ and R.U.\ contributed to the conception and design of the work, data acquisition and curation, technical implementation and evaluation framework.
C.B., K.Y.\ and R.L.\ provided clinical inputs to the study.
T.N.\ and H.P.\ supervised the work.
All authors contributed to the drafting and revision of the manuscript.

\section*{Ethics}\label{ethics}
This study was approved by the Providence Institutional Review Board, with a waiver of informed consent owing to the retrospective analysis of de-identified data.
The MSK-CHORD cohort is a publicly available, de-identified dataset.
There was no patient or public involvement in the design or conduct of this study. No formal study protocol was prepared and the study was not registered, as this is a retrospective observational analysis of de-identified data.

\section*{Competing Interests}\label{competing_interests}
J.M.Z.C., P.A., R.U., T.N.\ and H.P.\ are employees of Microsoft and may own stock in the company.
C.B.\ is a member of the scientific advisory board and owns stock in PrimeVax and BioAI; is on the scientific board of Lunaphore and SironaDx; has a consultant or advisory relationship with Sanofi, Agilent, Roche and Incendia; contributes to institutional research for Illumina; and is an inventor on US patent applications US20180322632A1 and US20200388033A1 filed by Providence Health and Services Oregon, Omics Data Automation.
R.L.\ is an inventor on US Patent US8415100B2 EsoGuard and EsoCheck devices with Lucid Diagnostics. KY has a sponsored research agreement with Bicara Therapeutics, material for research provided by Corbus Pharmaceuticals, and is an inventor on US Patents US-12560591-B2, US-20260042817-A1, and US-20170239283-A1.

\end{document}